\documentclass[conference]{IEEEtran}
\IEEEoverridecommandlockouts
\usepackage{cite}
\usepackage{amsmath,amssymb,amsfonts}
\usepackage{algorithmic}
\usepackage{array}    % 用于调整列宽和对齐
\usepackage{subcaption} % 在导言区加载 subcaption 宏包
\usepackage{multirow} % 用于合并单元格
\usepackage{graphicx}
\usepackage{booktabs} % 用于三线表
\usepackage{textcomp}
\usepackage{xcolor}
\usepackage[labelformat=simple]{subcaption}
\def\BibTeX{{\rm B\kern-.05em{\sc i\kern-.025em b}\kern-.08em
    T\kern-.1667em\lower.7ex\hbox{E}\kern-.125emX}}
\begin{document}

\title{Large Language Models for Single-Step and Multi-Step Flight Trajectory Prediction}

\author{\IEEEauthorblockN{Kaiwei Luo}
\IEEEauthorblockA{\textit{School of Computer Science} \\
\textit{Sichuan University}\\
Chengdu, China \\
kevin\_rowe\_it\_sc@stu.scu.edu.cn}
\and
\IEEEauthorblockN{Jiliu Zhou\textsuperscript{*}}
\IEEEauthorblockA{\textit{School of Computer Science} \\
\textit{Sichuan University}\\
Chengdu, China \\
zhoujl@cuit.edu.cn}
\thanks{\textsuperscript{*}Corresponding author.}
}
\maketitle

\begin{abstract}
Flight trajectory prediction is a critical time series task in aviation. While deep learning methods have shown significant promise, the application of large language models (LLMs) to this domain remains underexplored. This study pioneers the use of LLMs for flight trajectory prediction by reframing it as a language modeling problem. Specifically, We extract features representing the aircraft's position and status from ADS-B flight data to construct a prompt-based dataset, where trajectory waypoints are converted into language tokens. The dataset is then employed to fine-tune LLMs, enabling them to learn complex spatiotemporal patterns for accurate predictions. Comprehensive experiments demonstrate that LLMs achieve notable performance improvements in both single-step and multi-step predictions compared to traditional methods, with LLaMA-3.1 model achieving the highest overall accuracy. However, the high inference latency of LLMs poses a challenge for real-time applications, underscoring the need for further research in this promising direction.
\end{abstract}


\section{Introduction}
Air Traffic Management (ATM) faces significant challenges arising from the increasing density of flight activities. The rapid growth of the global economy has significantly boosted the demand for air transportation across various industries, causing higher airspace complexity\cite{Zhang2023}. To tackle these challenges, substantial global efforts have been made to develop more efficient air traffic systems. For instance, the United States has introduced the Next Generation Air Transportation System (NextGEN)\cite{Strohmeier2014} to modernize the national airspace, while Europe has launched the Single European Sky ATM Research (SESAR)\cite{Bolic2021} program to optimize air traffic management across member states. Both initiatives rely on Trajectory-Based Operations (TBO) for ATM automation\cite{Cate2013}. As an essential component of TBO, flight trajectory prediction systems ensure accurate and timely support in many ATM scenarios, such as conflict detection\cite{Chen2020}, flight delay forecasting\cite{Mamdouh2024}, and air traffic flow management\cite{ZhangY2024}.

\begin{figure}[tp]
\centerline{\includegraphics[width=\columnwidth]{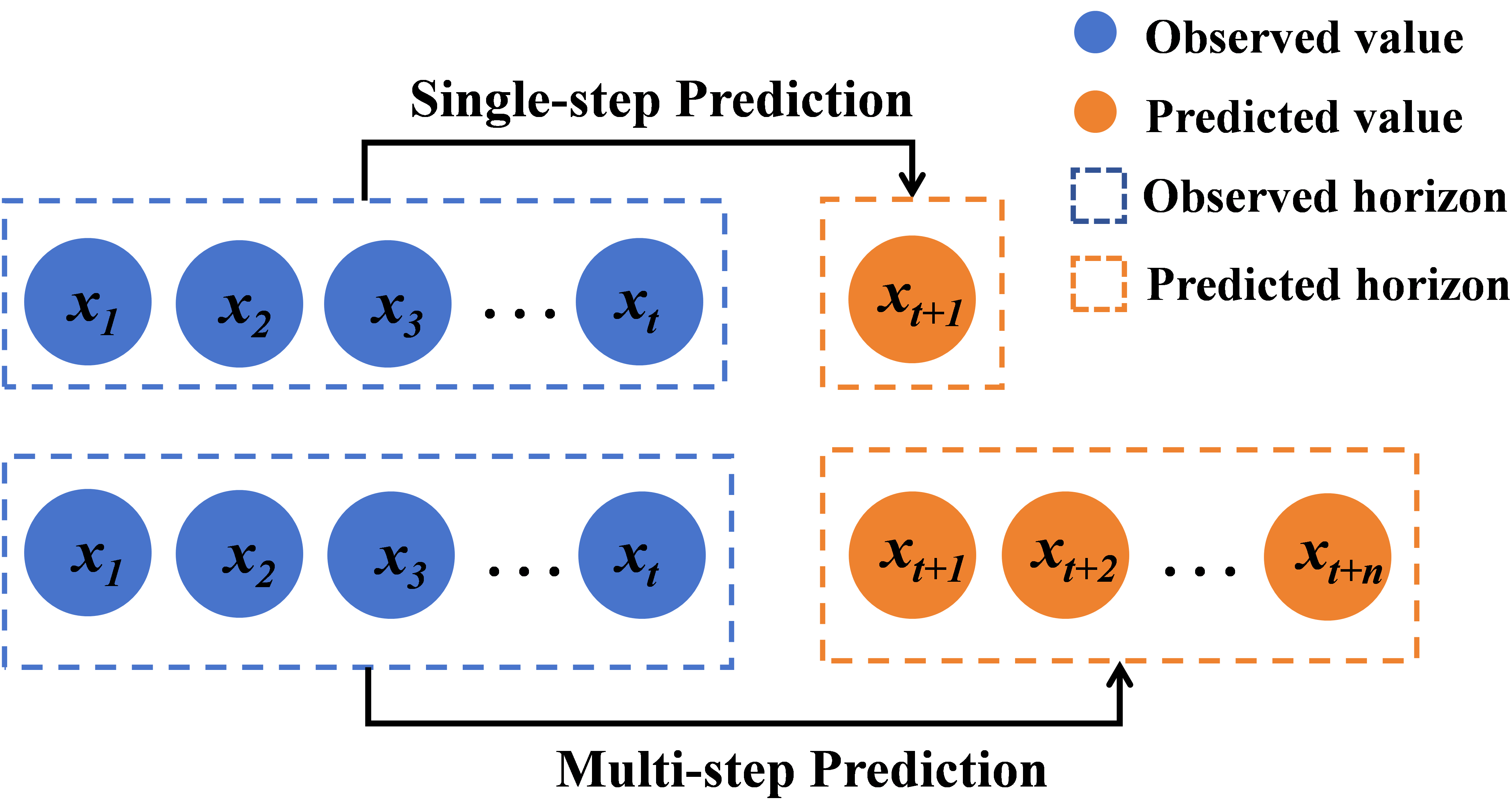}}
\caption{Overview of single-step and multi-step prediction in time series tasks.}
\label{prediction}
\vspace{-10pt} 
\end{figure}

Flight trajectory prediction is often viewed as a multivariate time series problem. Depending on the prediction horizon, time series tasks are categorized into single-step and multi-step prediction, as shown in Fig.~\ref{prediction}. In single-step prediction, the model predicts only the next immediate value, while in multi-step prediction, it forecasts multiple future values in one go. The primary goal of trajectory prediction is to forecast the future status parameters of aircraft, such as longitude, latitude, altitude, and velocity, based on the observed historical data\cite{Guo2024}. Trajectory prediction can be further divided into short-term and long-term categories based on the time scale\cite{Wang2017}. Short-term prediction emphasizes real-time responsiveness in dynamic environments, providing high-precision position estimates over short periods. In contrast, long-term prediction offers a broader perspective for strategic planning by incorporating external factors such as flight intentions and environmental data, but suffers from growing uncertainty and computational overhead. Therefore, this work focuses on short-term prediction in both single-step and multi-step scenarios, relying exclusively on historical status parameters.

\begin{figure}[tp]
\centerline{\includegraphics[width=\columnwidth]{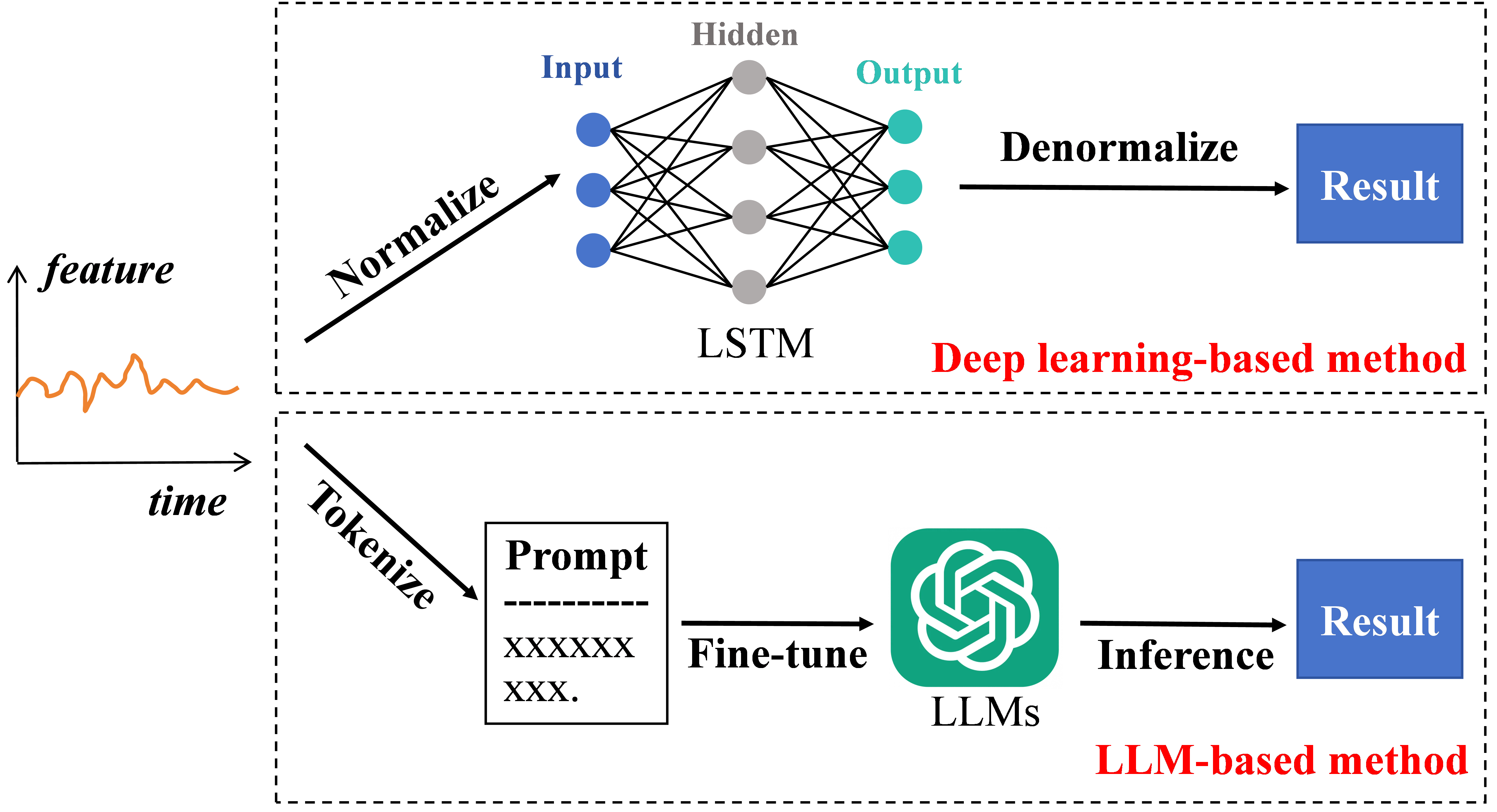}} 
\caption{Comparison of deep learning-based (e.g. LSTM) and LLM-based methods in time series tasks.}
\label{comparison}
\vspace{-10pt} 
\end{figure}

Great efforts have been made in flight trajectory prediction, evolving from physics-based\cite{Taob2014},\cite{Rezaie2018},\cite{Schuster2015} to data-driven approaches\cite{Zeng2020},\cite{Fan2024},\cite{Ma2020}. Physics-based methods typically model the interactions between the aircraft and environment using aerodynamics and kinematic equations. However, these models are too idealized to make accurate predictions in dynamic real-time air traffic systems\cite{Guo2023}. With the rise of deep learning, new avenues have been opened for prediction tasks. Deep learning models effectively capture complex air traffic dynamics from historical flight data, making them the most prevailing approach for this task. In recent years, large language models (LLMs) have advanced rapidly and been successfully applied to various areas, including computer vision\cite{Zhu2023},\cite{VisualPrompt2024}, speech recognition\cite{Fathullah2024}, and autonomous driving\cite{Mao2023}, demonstrating exceptional potential in solving complex challenges. The comparison of deep learning-based methods and LLM-based methods is detailed in Fig.~\ref{comparison}. 

As shown in Fig.~\ref{comparison}, neural network (e.g., LSTM\cite{Shi2018}), along with its variants, is widely used in deep learning-based methods, where data is normalized to unify feature scales. However, normalization introduces drawbacks as well, such as diluted data distribution, reduced data discrimination, and unexpected output when the input is out of range. LLM-based methods significantly reduce the reliance on explicit normalization by adopting a structured workflow comprising tokenization, prompt construction, fine-tuning, and inference.

While LLMs have demonstrated remarkable success across multiple domains, their potential in flight trajectory prediction remains insufficiently explored. To this end, this study investigates the capabilities of LLMs in trajectory prediction. Specifically, we extract relevant features representing flight status from Automatic Dependent Surveillance-Broadcast (ADS-B)\cite{CollisionFree2025} data and incorporate them into domain-specific prompts. We then apply Parameter-Efficient Fine-Tuning (PEFT) to various open-source LLMs, enabling them to learn underlying patterns from historical flight data. Finally, we predict future trajectories using the fine-tuned LLMs. Our contributions can be summarized as follows:
\begin{itemize}
\item We propose FTP-LLM (Large Language Models for Flight Trajectory Prediction), a novel framework that reformulate the prediction task as a language modeling problem. To the best of our knowledge, this is the first study to apply LLMs to flight trajectory prediction.
\item We construct datasets based on ADS-B flight data, and design aviation domain-specific prompt templates tailored for single-step and multi-step trajectory prediction.
\item We conduct extensive experiments to evaluate eight state-of-the-art LLMs, showcasing their strong performance and notable few-shot generalization capabilities in flight trajectory prediction.
\end{itemize}
The remainder of this paper is organized as follows: Section II provides a review of related work on flight trajectory prediction; Section III details the proposed methodology and model architecture; Section IV presents experimental results, analysis, and visualizations; and Section V concludes with a discussion of future research directions.

\section{Related Work}
\subsection{Flight Trajectory Prediction}
Existing approaches in flight trajectory prediction are classified into state estimation, kinetic, and data-driven methods.
\subsubsection{State Estimation Methods}
State estimation methods regard trajectory prediction as a mathematical state transition problem, focusing on dynamic state parameters like position, velocity, and acceleration. Kalman Filter (KF) and Hidden Markov Model (HMM) are two widely used state estimation techniques: KF is efficient for linear systems, while HMM excels in modeling discrete state transitions. Jeung et al.\cite{Jeung2007} employed HMMs to overcome the limitations of traditional space-partitioning methods in trajectory pattern mining. Wang et al.\cite{Taob2014} enhanced 4D trajectory prediction with a real-time noise-adaptive Kalman Filter. Rezaie et al.\cite{Rezaie2018} introduced a conditionally Markov sequence to model airliner trajectories using waypoint data. Despite their effectiveness, state estimation models struggle with nonlinear dynamics in the real-time systems due to simplified equations.
\subsubsection{Kinetic Methods}
Kinetic methods predict flight trajectories by modeling the relationships between forces and aircraft motion using differential equations. These models incorporate the aircraft's current state, meteorological conditions, and flight intent\cite{Zeng2022}. Schuster et al.\cite{Schuster2015} developed a 4D gate-to-gate model leveraging aircraft state and intent to enhance accuracy and air traffic management. Sun et al.\cite{Sun2016} inferred aircraft takeoff mass using a kinetic model and recursive runway motion data estimation. Besada et al.\cite{Besada2013} introduced intent-based formal languages and a trajectory processing engine for automated, hierarchical trajectory computation. In summary, kinetic methods rely on idealized assumptions, thus overlooking real-world constraints and human factors. Additionally, their reliance on extensive external data undoubtedly increases the computational intensity.
\begin{figure*}[htp]
\centerline{\includegraphics[width=\linewidth, height=7cm]{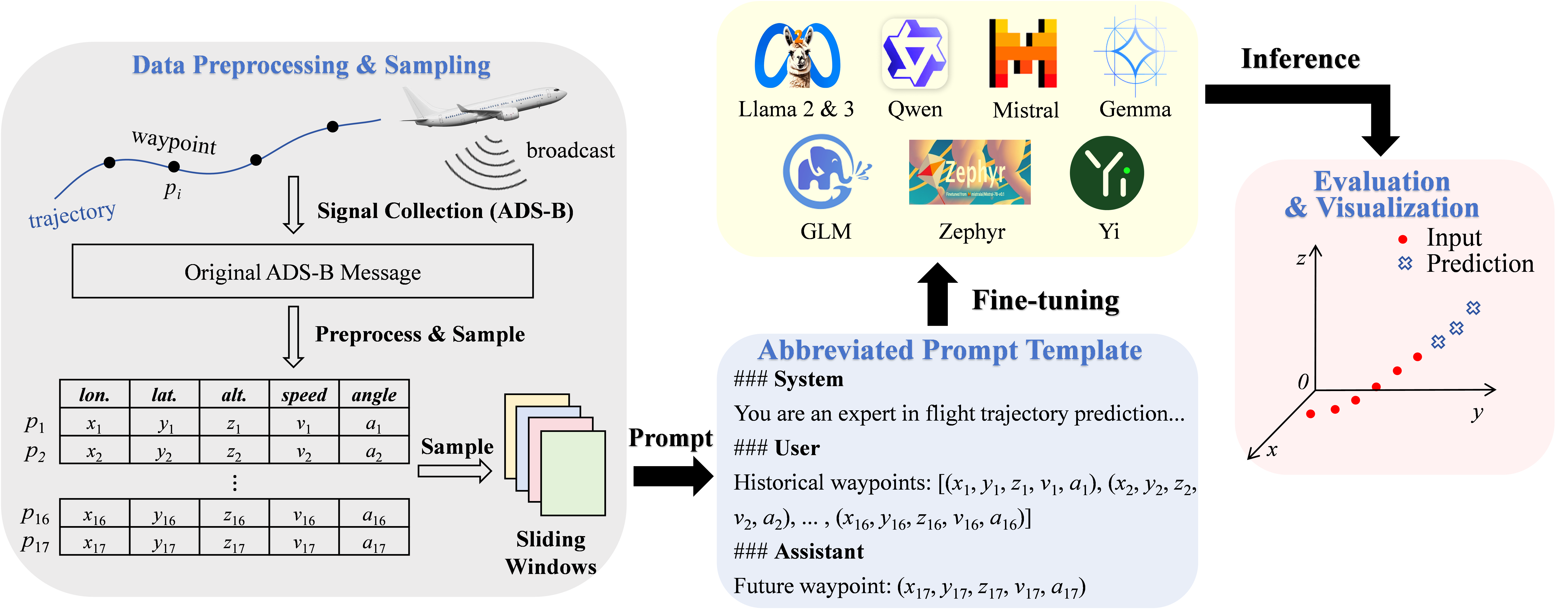}}
\caption{Overall architecture of the proposed FTP-LLM, comprising data preprocessing, sliding window sampling, prompt construction, fine-tuning on LLMs, and inference for prediction.}
\label{architecture} 
\vspace{-6pt}
\end{figure*}
\subsubsection{Data-Driven Methods}
Data-driven methods, which are primarily classified into machine learning and deep learning models, have attracted significant attention for their ability to directly learn patterns from data. Tastambekov et al.\cite{Tastambekov2014} introduced an algorithm employing local linear functional regression, integrating wavelet decomposition for data preprocessing and regression, to predict 4D short- to mid-term aircraft trajectories. De Leege et al.\cite{DeLeege2013} developed a machine learning approach that leverages historical trajectory and meteorological data to enhance the accuracy of aircraft trajectory prediction. Deep neural networks have become dominant tools across various domains, excelling in capturing complex relationships and efficiently handling large-scale datasets. Given the daily consistency of flight plans, Long Short-Term Memory (LSTM)\cite{Hochreiter1997} networks are particularly well-suited for capturing long-term dependencies and patterns in time series flight data. Shi et al.\cite{Shi2018} introduced an LSTM-based trajectory prediction model, achieving better accuracy compared to traditional models and establishing a robust foundation for anomaly detection and decision-making. Ma et al.\cite{Ma2020} developed a hybrid CNN-LSTM model for aircraft trajectory prediction, where the CNN extracts spatial features from adjacent trajectory regions, and the LSTM captures temporal dependencies. 

Originally proposed for machine translation, the Transformer\cite{Vaswani2017} model has revolutionized deep learning through its innovative multi-head attention mechanism. Guo et al.\cite{Guo2023} proposed FlightBERT, a Transformer-based framework for trajectory prediction, leveraging binary encoding and attribute correlation attention to capture complex motion patterns. Dong et al.\cite{Dong2024} employed the Transformer network to develop a comprehensive trajectory prediction model that spans the entire flight phase, from takeoff to landing. Fan et al.\cite{Dong2023} proposed a TCN-Informer model for aviation trajectory prediction, achieving high accuracy through spatiotemporal feature extraction and efficient temporal correlation.
\subsection{Large Language Models}
With the growing popularity and widespread application of Generative Pre-trained Transformers (GPT)\cite{Radford2018}, LLMs have emerged as leading tools for various domains. Extensive research has been carried out on time series tasks utilizing LLMs. Chang et al.\cite{Chang2023} proposed LLM4TS, leveraging pre-trained LLMs for time-series forecasting through a two-stage fine-tuning process and PEFT techniques. Munir et al.\cite{Munir2024} explored the feasibility of open-source LLMs for the ego-vehicle trajectory prediction problem in autonomous driving. Zhang et al.\cite{Zhang2024} applied the LLaMA model to flight trajectory reconstruction, demonstrating the efficiency of LLMs in handling noisy flight data but highlighting their limitations with long sequences due to token length constraints. Liu et al.\cite{Liu2024} pioneered the use of LLMs for cuffless blood pressure estimation from wearable biosignals through context-enhanced prompts and instruction tuning.
\section{Methodology}
In this section, we present FTP-LLM, a framework for trajectory prediction, as illustrated in Fig.~\ref{architecture}. We begin by explaining how LLMs can be applied to trajectory prediction, followed by a detailed description of the model architecture.
\subsection{Problem Definition}
Our objective is to predict an aircraft's position over several successive time steps based on historical data. Specifically, the flight trajectory $T$ is discretized into a sequence of waypoints:
\begin{equation}
T = \{T_{1:t}, T_{t+1:t+n}\},
\end{equation}
\begin{equation}
T_{1:t} = \{p_1, p_2, \dots, p_t\},
\end{equation}
\begin{equation}
T_{t+1:t+n} = \{p_{t+1}, p_{t+2}, \dots, p_{t+n}\},
\end{equation}
where a complete trajectory $T$ is divided into two parts: $T_{1:t}$, representing the $t$ previous waypoints, and $T_{t+1:t+n}$, denoting the $n$ future waypoints. Each waypoint $p_i$ captures the aircraft’s position and states at timestamp $i$, described by five attributes:
\begin{equation}
p_i = (x_i, y_i, z_i, v_i, a_i),
\end{equation}
where $x_i$, $y_i$, $z_i$, $v_i$, and $a_i$ correspond to longitude, latitude, altitude, speed, and heading angle, respectively. 

LLMs are designed to process language inputs, while trajectory waypoints are numerical coordinates. To bridge this gap, we embed these coordinates into prompts and use an LLM tokenizer to convert the trajectory $T$ into a sequence of language tokens:
\begin{equation}
T_{1:t} = \{p_1, p_2, \dots, p_t\} = \{w_1, w_2, \dots, w_n\},
\end{equation}
where $w_j$ denotes the $j$-th token in a sentence. Typically, waypoint $p_i$ is represented by a set of $w_j$. For instance, the longitude value “103.25” is split into three distinct tokens: “103”, “.”, and “25” using the LLaMA-3.1 tokenizer. In this way, trajectory prediction can be viewed as a next token prediction process and treated as a language modeling problem:
\begin{equation}
\mathcal{L}_{\text{LLM}} = -\sum_{j=1}^N \log P(\hat{w}_j \mid w_1, w_2, \dots, w_{j-1}).
\end{equation}

By performing data-to-tokens conversion on trajectory, we can then leverage LLMs to solve forecasting tasks. After decorating the trajectory with domain-specific prompt and fine-tuning LLMs on large-scale data, they can make trajectory predictions based on the learned probability distribution.

\subsection{Model Architecture}
\subsubsection{Data Preprocessing}
We collected flight trajectories of inbound and outbound, domestic and international flights at Guangzhou Baiyun International Airport (CAN), Beijing Capital International Airport (PEK), and Shanghai Pudong International Airport (PVG) to construct the datasets. A trajectory consists of multiple waypoints, each represented in ADS-B format and containing attributes such as timestamp, UTC time, callsign, longitude, latitude, altitude, velocity, and heading angle. Table~\ref{ADS-B} shows an example of ADS-B data. Incomplete, duplicate, and invalid trajectories (e.g., those with out-of-range latitude values) are first removed during preprocessing. Next, the values of longitude, latitude, altitude, velocity, and heading angle are rounded to 5, 5, 3, 3, and 2 decimal places, respectively. The raw ADS-B data, recorded at irregular intervals in seconds, are aggregated at the minute level to ensure temporal consistency. Specifically, for each interval, we realign the time unit by computing the average of longitude, latitude, altitude, velocity, and heading angle, respectively.
\begin{table}[bp]
\vspace{-6pt}
\caption{Features of ADS-B Data} % 表格标题
\centering
\begin{tabular}{ccc} % 不使用竖线，保持三线表简洁
\toprule % 表格顶部的横线
\textbf{Feature} & \textbf{Unit} & \textbf{Waypoint} \\
\midrule % 表格中间的横线
Timestamp      & Unix              & 1727926166              \\
UTC Time       & /                 & 2024-10-3 3:29:26       \\
Call Sign      & /                 & 3S528                   \\
Longitude      & Degree            & 13.61184                \\
Latitude       & Degree            & 50.48944                \\
Altitude       & Meter             & 10058.400               \\
Velocity       & Kilometer/Hour    & 968.596                 \\
Heading Angle  & Degree            & 125.00                  \\
\bottomrule % 表格底部的横线
\end{tabular}
\label{ADS-B}
\vspace{-10pt}
\end{table}

\begin{figure}[t]
\centering
\includegraphics[width=\linewidth, height=6.5cm]{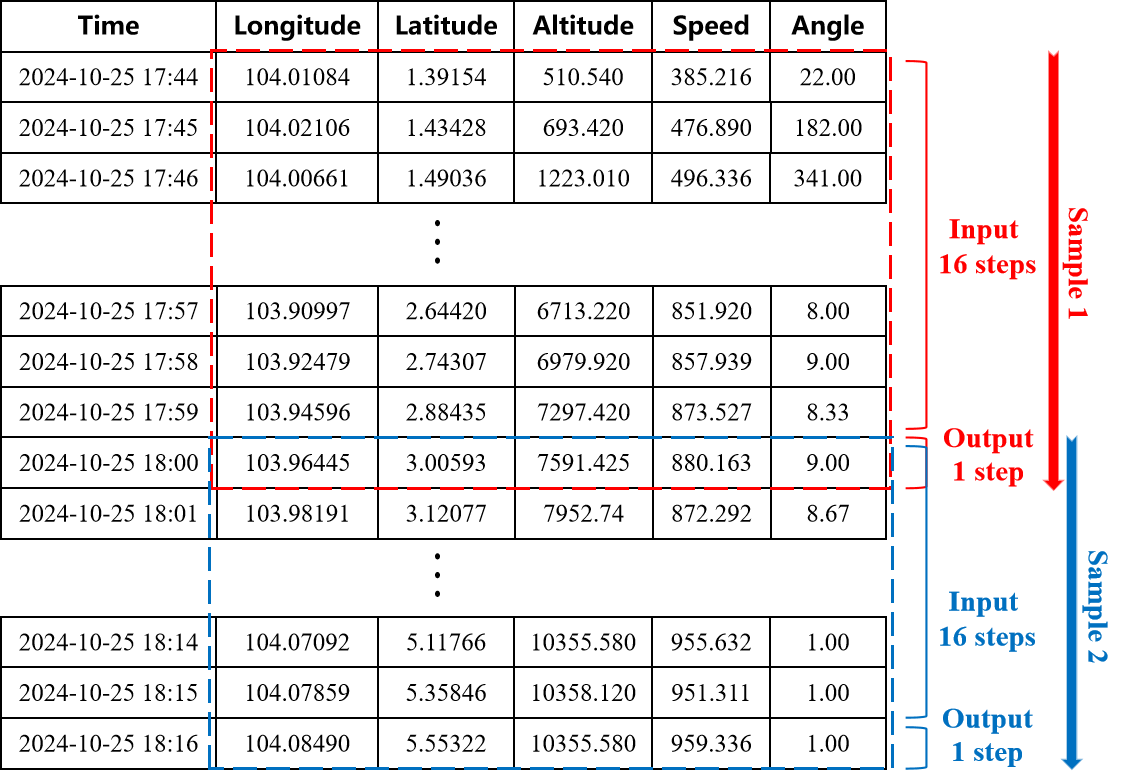} % 调整为适合宽度的比例
\caption{Visualization of the sliding window strategy in single-step prediction.}
\label{sliding}
\vspace{-6pt} 
\end{figure}
\subsubsection{Sliding Window Sampling}
A sliding window strategy is used to slice trajectories from the ADS-B data. To ensure continuity, the time interval within each window is strictly constrained to 1 minute, as aircraft may make stopovers during flight, leading to intermittent trajectory. Fig.~\ref{sliding} visualizes the sliding window strategy in detail. As shown in Fig.~\ref{sliding}, the window size is set to 17 for single-step prediction, consisting of 16 consecutive time steps as input and 1 time step for prediction. For multi-step prediction, the window size is set to 20 or 24, corresponding to 4 or 8 prediction steps, respectively. The stride is set to be larger than the window size to prevent overlap and enhance data diversity.
\subsubsection{Prompt Design}
\begin{figure}[t]
\centering
\includegraphics[width=\linewidth, height=7cm]{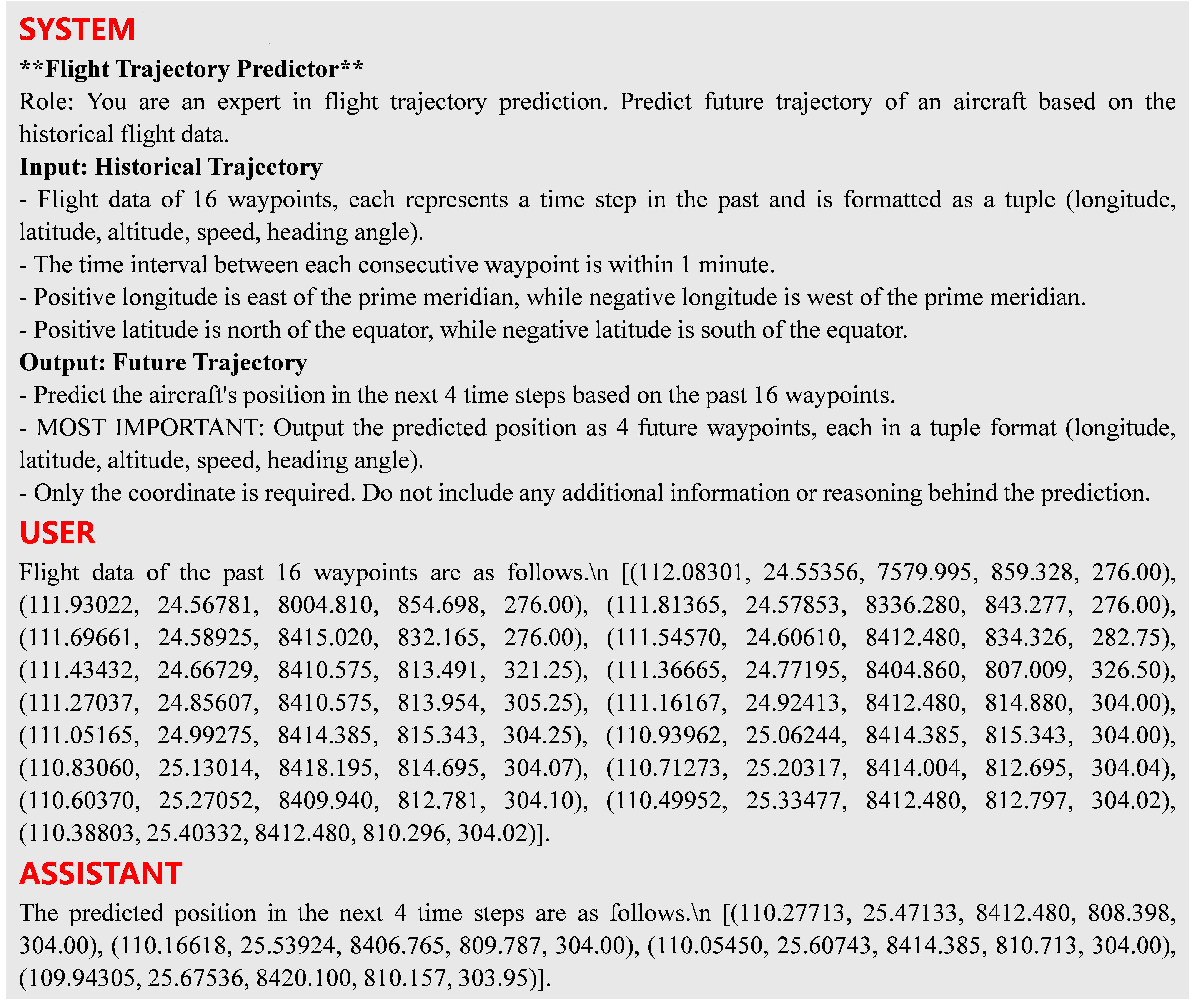} % 调整为适合宽度的比例
\caption{An example of prompt template provided to LLMs.}
\label{prompt}
\vspace{-10pt} 
\end{figure}

After processing and sampling the original ADS-B data, we populate the prompts with values of sliding windows. A typical chat-LLM prompt consists of three parts: system, user, and assistant. The system part elaborate the background and requirements of the flight trajectory prediction task, beginning with a role definition such as “You are an expert in flight prediction” to guide the model into the aviation domain. It also includes necessary terminology explanations and specifies output priorities. The user part provides waypoints from the previous 16 time steps in coordinate format to form a query. Finally, the assistant part contains the predicted waypoints for the next few time steps as the model’s response. During the fine-tuning phase, all three parts are incorporated into a single input using a specific tokenizer template. However, in the inference phase, the assistant part is masked, and the model generates predictions based only on the system and user parts. An example of prompts provided to LLMs is shown in Fig.~\ref{prompt}.

\section{Experiments}
\subsection{Experimental Configurations}
We conducted experiments on eight state-of-the-art open-source LLMs with parameters around 7 billion from HuggingFace, including Gemma-2-9B\cite{Team2024}, GLM-4-9B\cite{GLM2024}, LLaMA-2-7B\cite{Touvron2023}, LLaMA-3.1-8B\cite{Grattafiori2024}, Mistral-7B-v0.2\cite{Jiang2023}, Qwen-2.5-7B\cite{Qwen2025}, Yi-1.5-9B\cite{01AI2024}, and Zephyr-7B-Beta\cite{Tunstall2023}. For comparison, baseline trajectory prediction methods in this study include vanilla LSTM\cite{Shi2018}, BiLSTM\cite{Schuster1997} and Transformer\cite{Vaswani2017}. To make a trade-off between model performance and computational efficiency, we adopted Low-Rank Adaptation (LORA)\cite{Hu2022}, a PEFT technique, combined with 4-bit quantization to reduce memory usage and computational overhead. The whole fine-tuning phase lasts for 3 epochs with the batch size set to 4 and the initial learning rate for Adam optimizer is set to 0.0002. Both fine-tuning and inference were executed on a single RTX 4090 GPU with 24 GB of memory.
\subsection{Evaluation Metrics}
Two commonly used metrics, Mean Absolute Error (MAE) and Root Mean Square Error (RMSE), are employed to evaluate the performance of models in flight trajectory prediction. MAE quantifies the average absolute error, indicating how closely the predicted values align with the ground truth. In contrast, RMSE measures the square root of the mean squared differences between the predicted values and the ground truth. Smaller MAE and RMSE values reflect higher accuracy in motion prediction. The mathematical definitions of MAE and RMSE are as follows:
\begin{equation}
\text{MAE} = \frac{1}{n} \sum_{i=1}^n |y - y'|,
\end{equation}
\begin{equation}
\text{RMSE} = \sqrt{\frac{1}{n} \sum_{i=1}^n (y - y')^2},
\end{equation}
where $y$ and $y'$ denote the predicted and ground truth values of waypoint attributes, respectively. Additionally, average inference latency is introduced as a metric to evaluate the efficiency of a model. It is defined as the time delay between receiving a prompt and generating a prediction.
\begin{equation}
\text{Inference Latency} = {\frac{1}{N} \sum_{i=1}^N t_i},
\end{equation}
where $N$ is the size of test dataset and $t_i$ is the inference latency for $i$-th sample in the dataset.

\subsection{Experimental Results}
\subsubsection{Comparative Analysis}
The results of different models are presented in Table~\ref{results}.
\begin{table*}[ht] % 使用 table* 实现跨两栏
\caption{Results of Different Models in Flight Trajectory Prediction}
\centering
\begin{tabular}{ccccccccc}
\toprule
\multirow{2}{*}{\textbf{Model}} & \multirow{2}{*}{\textbf{Prediction}} & \multirow{2}{*}{\textbf{Inference}} & \multicolumn{3}{c}{\textbf{MAE ↓}}      & \multicolumn{3}{c}{\textbf{RMSE ↓}}     \\
\cmidrule(lr){4-6} \cmidrule(lr){7-9}
& \textbf{Steps} & \textbf{Latency (s)}& longitude (°)& latitude (°)& altitude (m)& longitude  (°)& latitude (°)& altitude  (m)\\
\midrule
\multirow{3}{*}{Gemma-2-9B\cite{Team2024}} 
    & 1 & 3.4534 & 0.0052 & 0.0044 & 22.3933 & \textbf{0.0097} & 0.0078 & 52.0669 \\ 
    & 4 & 7.9925 & 0.0208 & 0.0173 & 78.2016 & 0.0448 & 0.0369 & 188.7158 \\
    & 8 & 13.5666 & 0.0448 & 0.0413 & 143.2044 & 0.0981 & 0.0928 & 340.5264 \\
\midrule
\multirow{3}{*}{GLM-4-9B\cite{GLM2024}}  
    & 1 & 1.6104 & 0.0056 & 0.0048 & 23.0343 & 0.0104 & 0.0087 & 52.7163 \\
    & 4 & 4.5394 & 0.0185 & 0.0154 & 73.2479 & 0.0418 & 0.0358 & 181.1112 \\
    & 8 & 8.2889 & 0.0472 & 0.0428 & 140.2281 & 0.1016 & 0.0941 & 341.4865 \\
\midrule
\multirow{3}{*}{LLaMA-2-7B\cite{Touvron2023}}  
    & 1 & 1.4704 & 0.0061 & 0.0051 & 23.7410 & 0.0114 & 0.0088 & 53.8996 \\
    & 4 & 4.3878 & 0.0186 & 0.0150 & 72.4074 & 0.0419 & 0.0329 & 178.9544 \\
    & 8 & 7.9162 & 0.0488 & 0.0445 & 147.4378 & 0.1039 & 0.0970 & 349.4891 \\
\midrule
\multirow{3}{*}{LLaMA-3.1-8B\cite{Grattafiori2024}}
    & 1 & 1.0585 & 0.0053 & 0.0046 & 23.2201 & 0.0098 & 0.0081 & 53.4848 \\
    & 4 & 4.2812 & 0.0169 & \textbf{0.0134} & \textbf{68.2341} & 0.0398 & \textbf{0.0304} & 174.7966 \\
    & 8 & 6.6732 & \textbf{0.0434} & \textbf{0.0403} & \textbf{138.5888} & \textbf{0.0955} & \textbf{0.0904} & \textbf{336.9702} \\
\midrule
\multirow{3}{*}{Mistral-7B-v0.2\cite{Jiang2023}} 
    & 1 & 2.1465 & \textbf{0.0051} & \textbf{0.0042} & \textbf{21.2091} & \textbf{0.0097} & \textbf{0.0077} & \textbf{49.9429} \\
    & 4 & 6.3397 & 0.0171 & 0.0142 & 70.9545 & 0.0398 & 0.0328 & 178.0764 \\
    & 8 & 9.4770 & 0.0457 & 0.0430 & 147.3300 & 0.1005 & 0.0975 & 347.3605 \\
\midrule
\multirow{3}{*}{Qwen-2.5-7B\cite{Qwen2025}}    
    & 1 & 1.8655 & 0.0055 & 0.0050 & 23.3936 & 0.0102 & 0.0171 & 53.6241 \\
    & 4 & 5.3395 & 0.0186 & 0.0152 & 73.4411 & 0.0427 & 0.0335 & 178.6826 \\
    & 8 & 10.2877 & 0.0477 & 0.0445 & 148.1366 & 0.1012 & 0.0986 & 344.7497 \\
\midrule
\multirow{3}{*}{Yi-1.5-9B\cite{01AI2024}} 
    & 1 & 2.1667 & 0.0055 & 0.0053 & 23.6459 & 0.0107 & 0.0190 & 53.3111 \\ 
    & 4 & 7.9780 & 0.0181 & 0.0149 & 71.3618 & 0.0417 & 0.0330 & 177.8078 \\
    & 8 & 12.4595 & 0.0465 & 0.0439 & 147.4845 & 0.1015 & 0.0985 & 350.9740 \\
\midrule
\multirow{3}{*}{Zephyr-7B-Beta\cite{Tunstall2023}}
    & 1 & 2.1434 & 0.0059 & 0.0054 & 24.1520 & 0.0113 & 0.0124 & 54.7890 \\
    & 4 & 6.7279 & \textbf{0.0166} & 0.0137 & 68.4934 & \textbf{0.0391} & 0.0314 & \textbf{173.6565} \\
    & 8 & 9.2350 & 0.0469 & 0.0441 & 153.6472 & 0.1017 & 0.0983 & 360.0954 \\
\midrule
\multirow{3}{*}{LSTM\cite{Shi2018}} 
    & 1 & \textbf{0.0006} & 0.0065 & 0.0060 & 27.1511 & 0.0116 & 0.0100 & 59.2235 \\ 
    & 4 & \textbf{0.0006} & 0.0201 & 0.0167 & 82.2635 & 0.0436 & 0.0368 & 185.5921 \\
    & 8 & \textbf{0.0008} & 0.0488 & 0.0460 & 158.6258 & 0.1038 & 0.0994 & 368.4862 \\  
\midrule
\multirow{3}{*}{BiLSTM\cite{Schuster1997}} 
    & 1 & 0.0006 & 0.0066 & 0.0062 & 27.6808 & 0.0119 & 0.0105 & 60.3736 \\ 
    & 4 & 0.0007 & 0.0203 & 0.0169 & 83.1023 & 0.0441 & 0.0370 & 187.6258 \\
    & 8 & 0.0009 & 0.0491 & 0.0457 & 160.1134 & 0.1045 & 0.1002 & 372.4689 \\
\midrule
\multirow{3}{*}{Transformer\cite{Vaswani2017}} 
    & 1 & 0.0007 & 0.0059 & 0.0055 & 26.0105 & 0.0108 & 0.0098 & 57.8672 \\ 
    & 4 & 0.0009 & 0.0190 & 0.0161 & 80.1597 & 0.0417 & 0.0351 & 184.3560 \\
    & 8 & 0.0010 & 0.0484 & 0.0447 & 156.2346 & 0.1017 & 0.0987 & 357.4765 \\
\bottomrule
\end{tabular}
\label{results}
\end{table*}
\begin{table*}[!ht] % 使用 table* 实现跨两栏
\caption{Results of the LLaMA-3.1 Model in Different Flight Phases}
\centering
\begin{tabular}{ccccccccc}
\toprule
\multirow{2}{*}{\textbf{Model}} & \multirow{2}{*}{\textbf{Prediction Steps}} & \multirow{2}{*}{\textbf{Phase}}   & \multicolumn{3}{c}{\textbf{MAE ↓}}      & \multicolumn{3}{c}{\textbf{RMSE ↓}}     \\
\cmidrule(lr){4-6} \cmidrule(lr){7-9}
&  &  & longitude (°)& latitude (°)& altitude (m)& longitude (°)& latitude (°)& altitude (m)\\
\midrule
\multirow{12}{*}{LLaMA-3.1-8B\cite{Grattafiori2024}} &
\multirow{4}{*}{1} & {Entire} & 0.0053 & 0.0046 & 23.2201 & 0.0098 & 0.0081 & 53.4848 \\
                &  & {Take-off} & 0.0052 & 0.0045 & 57.5784 & 0.0092 & 0.0078 & 87.0801 \\
                &  & {Cruise} & 0.0058 & 0.0047 & 16.7268 & 0.0103 & 0.0082 & 47.1502 \\
                &  & {Landing} & 0.0029 & 0.0042 & 26.7520 & 0.0066 & 0.0078 & 44.7281 \\           
\cmidrule(lr){2-9}
& \multirow{4}{*}{4} & {Entire} & 0.0169 & 0.0134 & 68.2341 & 0.0398 & 0.0304 & 174.7966 \\
                  &  & {Take-off} & 0.0196 & 0.0161 & 169.8109 & 0.0415 & 0.0328 & 282.7475 \\
                  &  & {Cruise} & 0.0179 & 0.0137 & 52.3356 & 0.0416 & 0.0315 & 160.0391 \\
                  &  & {Landing} & 0.0106 & 0.0098 & 70.6739 & 0.0282 & 0.0221 & 136.0483 \\
\cmidrule(lr){2-9}
& \multirow{4}{*}{8} & {Entire} & 0.0434 & 0.0403 & 138.5888 & 0.0955 & 0.0904 & 336.9702 \\
                  &  & {Take-off} & 0.0553 & 0.0638 & 351.9017 & 0.1104 & 0.1218 & 571.8308 \\
                  &  & {Cruise} & 0.0500 & 0.0439 & 86.0120 & 0.1084 & 0.0956 & 275.0546 \\
                  &  & {Landing} & 0.0389 & 0.0337 & 149.5462 & 0.0881 & 0.0794 & 254.8113 \\
\bottomrule
\end{tabular}
\label{tab3}
\end{table*}
It is evident that LLMs outperform traditional deep learning methods in both single-step and multi-step prediction tasks. Among the LLMs, the Mistral-v0.2 model demonstrates the best performance in single-step prediction. For multi-step prediction, both LLaMA-3.1 and Zephyr-Beta models exhibit outstanding performance in the 4-step prediction task, with each achieving the best results in different metrics. However, in the 8-step prediction task, the LLaMA-3.1  surpasses all other models, achieving the lowest MAE and RMSE. For traditional deep learning models, the Transformer outperforms LSTM and BiLSTM in both single-step and multi-step prediction tasks, primarily due to its multi-head self-attention mechanism, which helps to capture global relationships within the data. Furthermore, the performance of LSTM and BiLSTM is comparable, indicating that the bidirectional structure of BiLSTM does not provide significant advantages in this task, where future context is less critical. A general performance degradation is observed across all models as the prediction horizon increases, which can be attributed to the growing challenge of accurately predicting successive trajectory waypoints in a single prediction and the lack of external knowledge. In addition, prediction errors for altitude are significantly larger than those for longitude and latitude. On the one hand, this discrepancy arises because altitude changes frequently and drastically during flight, whereas longitude and latitude typically exhibit only minor variations. On the other hand, altitude values, ranging from thousands to tens of thousands, differ substantially in magnitude compared to longitude (-180° to 180°) and latitude (-90° to -90°).

While LLMs achieve competitive results in terms of MAE and RMSE, they suffer from high inference latency compared to traditional deep learning models. This is mainly due to their complex architectures as well as the massive number of parameters, which necessitate substantial computational resources. Among all LLMs, the LLaMA-3.1 model stands out for its relatively lower latency thanks to its distinctive tokenizer, which treats numerical values (e.g., “123”) as a single token rather than splitting them into individual tokens (e.g., “1”, “2”, “3”). Consequently, LLaMA-3.1 generates fewer tokens during inference and reduces the overall inference time.
\subsubsection{Flight Phase-based Analysis}
This study further evaluates the predictive performance of the LLaMA-3.1 model in different flight phases. Flight trajectory is usually divided into three primary phases: take-off, cruise, and landing. Obvious differences in prediction performance are observed in different phases, as shown in Table~\ref{tab3}.

Specifically, in terms of MAE and RMSE metrics for longitude and latitude, the landing phase achieves the highest accuracy, followed by the cruise phase, with the take-off phase exhibiting the largest errors. The reasons can be summarized as follows: the landing phase typically adheres to regulated descent paths, resulting in smoother trajectories. The cruise phase experiences occasional influences from air currents and route adjustments, causing slightly higher errors in longitude and latitude. The take-off phase involves rapid acceleration and steep climbs, making it the most challenging phase for accurate predictions.

However, when it comes to the MAE and RMSE metrics for altitude, the cruise phase demonstrates the best performance, while the landing and take-off phases exhibit relatively larger errors. This is because the aircraft operates more stably with minimal altitude fluctuations during cruise, enabling the model to better capture underlying patterns. In contrast, the dynamic and nonlinear altitude changes during the landing and take-off phases introduce irregularity, increasing the complexity of making predictions. These findings underscore the need for models that incorporate the unique characteristics of each flight phase and encourage further research on phase-specific algorithms to enhance accuracy.

\begin{figure*}[ht]
    \centering
    % 第一行子图
    \begin{subfigure}[b]{0.3\textwidth}
        \includegraphics[width=\textwidth]{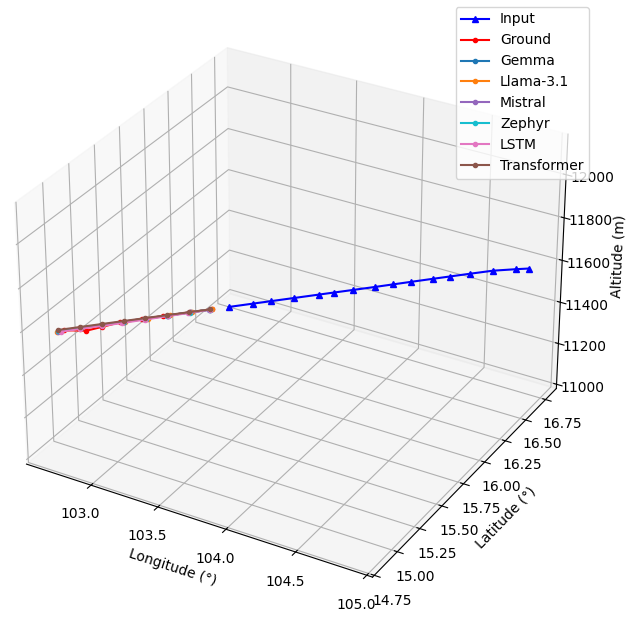} % 替换为图片路径
        \caption{Cruise phase.} % 子图标题
        \label{subfig41} % 子图引用标签
    \end{subfigure}
    \hspace{1em} % 子图之间的水平间距
    \begin{subfigure}[b]{0.3\textwidth}
        \includegraphics[width=\textwidth]{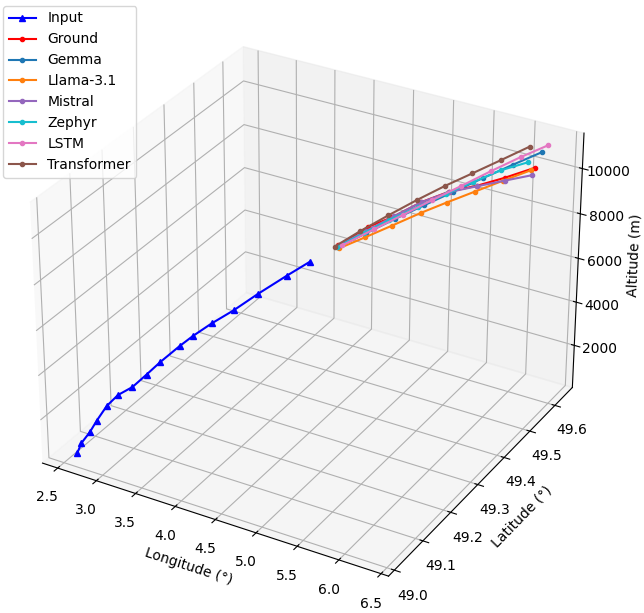}
        \caption{Take-off phase.}
        \label{subfig42} % 子图引用标签
    \end{subfigure}
    \hspace{1em}
    \begin{subfigure}[b]{0.3\textwidth}
        \includegraphics[width=\textwidth]{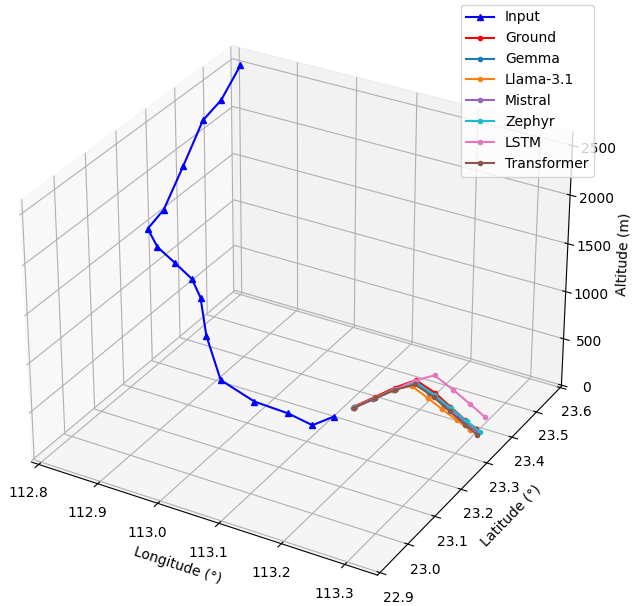}
        \caption{Landing phase.}
        \label{subfig43} % 子图引用标签
    \end{subfigure}

    % 第二行子图
    \vspace{1em} % 第一行与第二行之间的垂直间距
    \begin{subfigure}[b]{0.3\textwidth}
        \includegraphics[width=\textwidth]{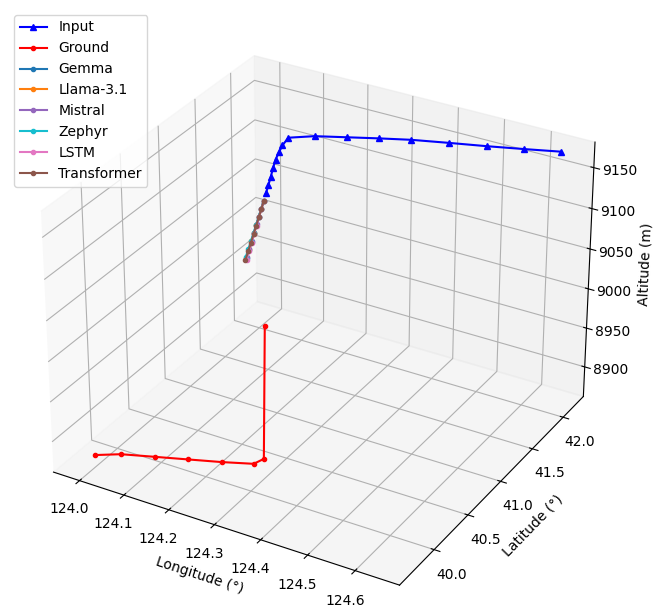}
        \caption{Cruise phase with sudden drop.}
        \label{subfig44} % 子图引用标签
    \end{subfigure}
    \hspace{1em}
    \begin{subfigure}[b]{0.3\textwidth}
        \includegraphics[width=\textwidth]{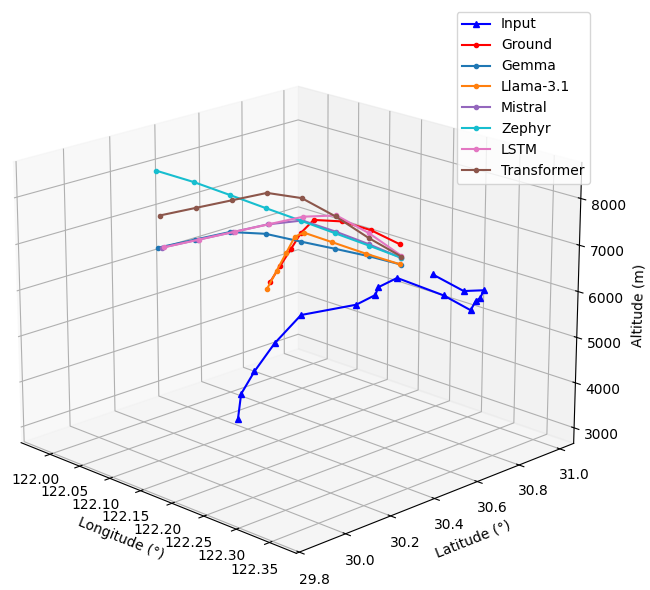}
        \caption{Take-off phase with slight turn.}
        \label{subfig45} % 子图引用标签
    \end{subfigure}
    \hspace{1em}
    \begin{subfigure}[b]{0.3\textwidth}
        \includegraphics[width=\textwidth]{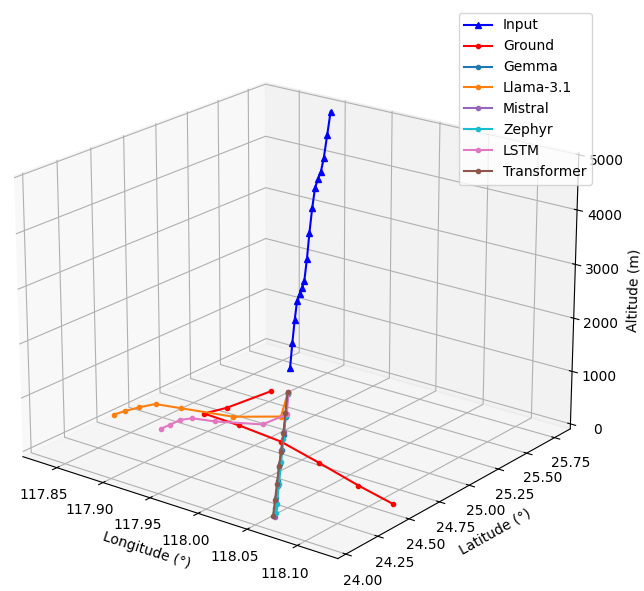}
        \caption{Landing phase with sharp turn.}
        \label{subfig46} % 子图引用标签
    \end{subfigure}

\caption{Visualization of 8-step prediction results in different flight phases. Blue for input, red for ground truth.}

\label{visualization} % 设置整个图的引用标签
\end{figure*}

\subsubsection{Visualization Analysis}
To provide an intuitive comparison, we visualize the prediction results of different models in 8-step predictions. Fig.~\ref{visualization} depicts representative scenarios in different flight phases. In Fig.~\ref{subfig41}, all models demonstrate accurate predictions, as the flight maintains a constant cruise altitude during this phase. In Fig.~\ref{subfig42} and Fig.~\ref{subfig43}, all models successfully capture the overall trajectory trends, with slight discrepancies observed between the predictions and the ground truth. The last three figures illustrate special cases encountered during flight. In Fig.~\ref{subfig44}, the flight initially maintains a stable altitude; however, an abrupt altitude drop of approximately 200 meters occurs at the first step of the prediction horizon, causing all models to fail in producing accurate predictions. Fig.~\ref{subfig45} presents a slight turning maneuver in the future trajectory, which is effectively captured by the LLaMA-3.1 model, while other models exhibit poorer performance. Finally, in Fig.~\ref{subfig46}, most models incorrectly predict that the aircraft will continue descending without turning. Unlike those models, the LLaMA-3.1 and Transformer models show some ability to recognize signs of a potential turn. However, they still can't make satisfactory predictions in the end.

It can be observed from theses figures that the predictive performance during the cruise phase is better than that during the take-off and landing phases. This is mainly due to the aircraft's stability during the cruise phase, which allows for more accurate trajectory predictions. However, unanticipated maneuvers during flight increase complexity that far exceeds the capabilities of fine-tuned LLMs, underscoring the need for further research.
\subsubsection{Few-shot Learning}
Furthermore, we investigated the generalization capability of LLMs, an essential aspect that distinguishes them from traditional deep learning methods. We conducted 4-step prediction experiments solely on the LLaMA-3.1 model, splitting the training data into different proportions: 1\%, 5\%, 10\%, 30\%, and 50\%. As presented in Table~\ref{tab4}, the results show that the LLaMA-3.1 model can still achieve satisfactory performance even with a limited amount of training data (approximately 30\%). This highlights its extensive pre-trained knowledge and powerful generalization as well as transfer learning abilities in few-shot learning scenarios. More importantly, these experiments offer the insight that, in contrast to traditional  deep learning-based models which typically demand tremendous training data, LL particularly well-suited for data-limited situations.
\begin{table*}[ht] % 使用 table* 实现跨两栏
\caption{Few-shot Learning Performance Between LLaMA-3.1 and Deep Learning Models}
\centering
\begin{tabular}{ccccccccc}
\toprule
\multirow{2}{*}{\textbf{Model}} & \multirow{2}{*}{\textbf{Proportion}}  & \multicolumn{3}{c}{\textbf{MAE ↓}}      & \multicolumn{3}{c}{\textbf{RMSE ↓}}     \\
\cmidrule(lr){3-5} \cmidrule(lr){6-8}
&  & longitude (°)& latitude (°)& altitude (m)& longitude (°)& latitude (°)& altitude (m)\\
\midrule
\multirow{5}{*}{LLaMA-3.1-8B\cite{Grattafiori2024}} 
    & 1 \%  & 0.0251 & 0.0214 & 98.8880 & 0.0500 & 0.0439 & 216.2459 \\ 
    & 5 \%  & 0.0230 & 0.0192 & 89.7103 & 0.0476 & 0.0387 & 201.7122 \\ 
    & 10 \%  & 0.0216 & 0.0176 & 84.1962 & 0.0471 & 0.0364 & 193.5335 \\
    & 30 \%  & 0.0200 & 0.0166 & 81.5059 & 0.0440 & 0.0349 & 192.2510 \\
    & 50 \%  & 0.0189 & 0.0155 & 77.2418 & 0.0423 & 0.0340 & 186.8130 \\
    & 100 \%  & 0.0169 & 0.0134 & 68.2341 & 0.0398 & 0.0304 & 174.7966 \\
\midrule
{LSTM\cite{Shi2018}} & 100 \% & 0.0201 & 0.0167 & 82.2635 & 0.0436 & 0.0368 & 185.5921 \\
\midrule
{BiLSTM\cite{Schuster1997}} & 100 \% & 0.0203 & 0.0169 & 83.1023 & 0.0441 & 0.0370 & 187.6258 \\
\midrule
{Transformer\cite{Vaswani2017}} & 100 \% & 0.0190 & 0.0161 & 80.1597 & 0.0417 & 0.0351 & 184.3560 \\
\midrule
\end{tabular}
\label{tab4}
\end{table*}
\begin{figure}[t] % 修改为放在页面顶部
\centering
\begin{minipage}[t]{0.48\textwidth} % 左栏占48%
    \centering
    \subcaptionbox{Missing trajectory.\label{subfig51}}{% 
        \includegraphics[width=\textwidth]{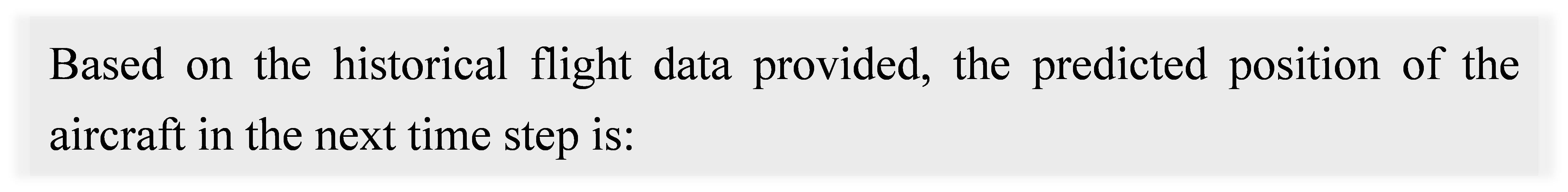}}\\ % 第一子图
    \vspace{0.1cm} % 子图间距
    \subcaptionbox{Unexpected format.\label{subfig52}}{% 
        \includegraphics[width=\textwidth]{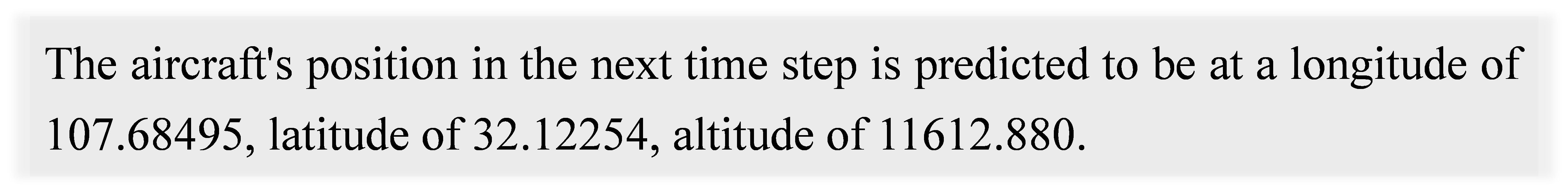}}\\ % 第二子图
    \vspace{0.1cm} % 子图间距
    \subcaptionbox{Severe deviation.\label{subfig53}}{% 
        \includegraphics[width=\textwidth]{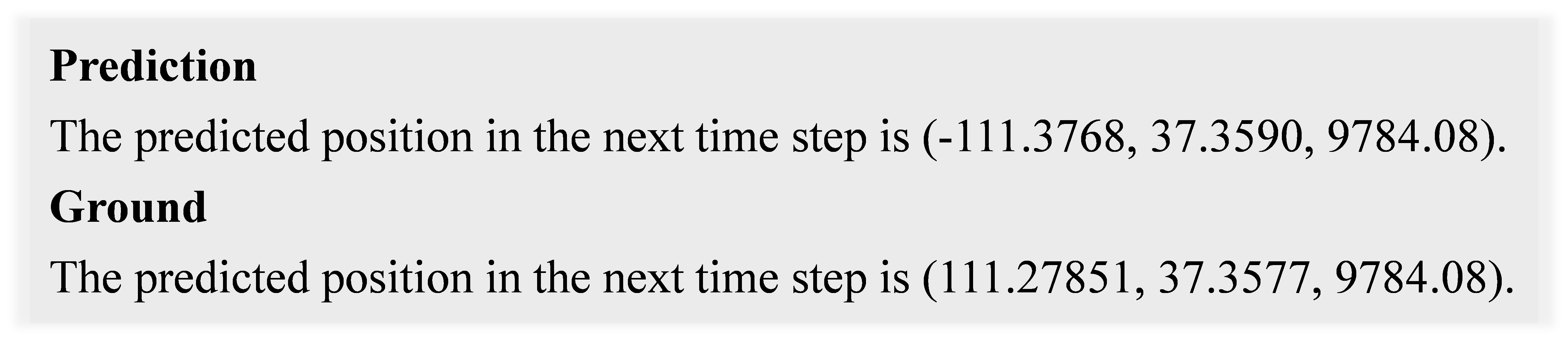}}   % 第三子图
\end{minipage}%
\begin{minipage}[t]{0.48\textwidth} % 右栏占48%
% 空白占位，确保左栏图片在左侧
\end{minipage}
\caption{Failure cases in the Yi-1.5 model.}
\vspace{-10pt} 
\label{failure}
\end{figure}

\subsubsection{Failure Analysis}
We observed a number of failure cases when using the Yi-1.5 model during inference, as illustrated in Fig.~\ref{failure}. In Fig.~\ref{subfig51}, the future trajectory is missing entirely. In Fig.~\ref{subfig52}, the output contains a complete trajectory, but is presented in an incorrect format, failing to represent it as coordinates. In Fig.~\ref{subfig53}, the prediction deviates significantly from the ground truth. The longitude value at the next time step is expected to be positive, yet the Yi-1.5 model outputs a negative value instead. This issue may result from insufficient fine-tuning of the model to comprehend the implications of a negative sign in longitude or latitude. To ensure the reliability of metrics, cases with severe deviations are excluded when calculating MAE and RMSE.

\section{Conclusion and future work}
In this paper, we pioneer the use of LLMs in flight trajectory prediction. Through comprehensive experiments on real ADS-B data, we demonstrated the potential of LLMs for both single-step and multi-step predictions compared to traditional deep learning-based methods. Besides, the visualization results showed that they 
can effectively understand and capture the underlying trajectory patterns across different phases. Moreover, generalization experiments on the LLaMA-3.1 model revealed that LLMs can make satisfactory predictions even with limited training data, highlighting their extensive pre-trained knowledge and strong transfer learning capability.

Even though LLMs exhibit strength in predicting future trajectory, their severe and unacceptable inference latency, especially as the prediction horizon extends, prevents them from meeting the requirements of real-time air traffic systems. To address this problem, inference acceleration techniques must be considered in future work. Regarding challenges, on the one hand, LLMs yield less accurate results when unexpected operations occur during flight, such as sudden drops or sharp turns. On the other hand, prediction errors vary significantly across different flight phases, emphasizing  the need for advanced algorithms tailored to each phase. Future research should focus on improving the robustness and accuracy of LLMs in flight trajectory prediction.

\section*{Acknowledgment} 
This work is supported in part by the National Natural Science Foundation of China (NSFC 62371325).

\bibliographystyle{IEEEtran}
\bibliography{LLM} 
\end{document}